# Survey on Deep Neural Networks in Speech and Vision Systems

M. Alam, M. D. Samad[1], L. Vidyaratne, A. Glandon and K. M. Iftekharuddin[*]

Vision Lab in Department of Electrical and Computer Engineering, Old Dominion University, Norfolk, VA 23529
(email: malam001, lvidy001, aglan001, kiftekha@odu.edu [*]corresponding author).
[1] Department of Computer Science, Tennessee State University, Nashville, TN, 37209 (email: msamad@tnstate.edu)

*Abstract*—This survey presents a review of state-of-the-art deep neural network architectures, algorithms, and systems in vision and speech applications. Recent advances in deep artificial neural network algorithms and architectures have spurred rapid innovation and development of intelligent vision and speech systems. With availability of vast amounts of sensor data and cloud computing for processing and training of deep neural networks, and with increased sophistication in mobile and embedded technology, the next-generation intelligent systems are poised to revolutionize personal and commercial computing. This survey begins by providing background and evolution of some of the most successful deep learning models for intelligent vision and speech systems to date. An overview of large-scale industrial research and development efforts is provided to emphasize future trends and prospects of intelligent vision and speech systems. Robust and efficient intelligent systems demand low-latency and high fidelity in resource-constrained hardware platforms such as mobile devices, robots, and automobiles. Therefore, this survey also provides a summary of key challenges and recent successes in running deep neural networks on hardware-restricted platforms, i.e. within limited memory, battery life, and processing capabilities. Finally, emerging applications of vision and speech across disciplines such as affective computing, intelligent transportation, and precision medicine are discussed. To our knowledge, this paper provides one of the most comprehensive surveys on the latest developments in intelligent vision and speech applications from the perspectives of both software and hardware systems. Many of these emerging technologies using deep neural networks show tremendous promise to revolutionize research and development for future vision and speech systems.

## 1. INTRODUCTION

THERE has been a massive accumulation of human-centric data to an unprecedented scale over the last two decades. This data explotion coupled with rapid growth in computing power have rejuvenated the field of neural networks and sophisticated intelligent system (IS). In the past, neural networks has mostly been limited to the application of industrial control and robotics. However, recent advancements in neural networks have led to successful applications of IS in almost every aspect of human life with the introduction of intelligent transportation [1-10], intelligent diagnosis and health monitoring for precision medicine [11-14], robotics and automation in home appliances [15], virtual online assistance [16], e-marketing [17], and weather forecasting and natural disasters monitoring [18] among others. The widespread success of IS technology has redefined and augmented human ability to communicate and comprehend the world by innovating on 'smart' physical systems. A 'smart' physical system is designed to interpret, act and collaborate with complex multimodal human senses such as vision, touch, speech, smell, gestures, or hearing. A large body of smart physical systems are developed targeting two primary senses used in human communication: vision and speech.

The advancement in speech and vision processing systems has enabled tremendous research and development in the areas of human-computer interactions [19], biometric applications [20, 21], security and surveillance [22], and most recently in computational behavioral analysis [23-27]. While traditional machine learning and evolutionary computations have enriched IS to solve complex pattern recognition problems over many decades, these techniques have limitations in their ability to process natural data or images in raw data formats. A number of computational steps are used to extract representative features from raw data or images prior to applying machine learning models. This intermediate representation of raw data, known as 'hand-engineered' features, requires domain expertise and human interpretation of physical patterns such as texture, shape, geometry, etc. There are three major problems with 'hand-engineered' features that impede major progress in IS. First, the choice of 'hand-engineered' features is application dependent and involves human interpretation and evaluation. Second, 'hand-engineered' features are extracted from each sample in a standalone manner without the knowledge of inevitable noise and variations in data. Third, 'hand-engineered' features may perform excellently with some input but may completely fail to extract quality features in other types of input data. This can lead to high variability in vision and speech recognition performance.

A solution to the limitations of 'hand-engineered' features has emerged through mimicking functions of biological neurons in artificial neural networks (ANN). The potential of ANNs is recently being exploited with access to large trainable datasets, efficient learning algorithms, and powerful computational resources. Few of these advancements in ANN over the last decade have led to deep learning [28, 29] that, in turn, has revolutionized several application domains including computer vision, speech analysis, biomedical image processing, and online market analyses. The rapid success of deep learning over traditional machine learning may be attributed to three factors. First, deep learning offers end-to-end trainable architectures that integrate feature extraction, dimensionality reduction, and final classification. These steps are otherwise treated as standalone sub-systems in conventional machine learning, which may result in suboptimal pattern recognition performance. Second, target-specific and informative features may be learned from both input examples and classification targets without resorting to application-specific feature extractors. Third, deep learning models are highly flexible in capturing complex nonlinear relationships between inputs and output targets at a level that is far beyond the capacity of 'hand-engineered' features.

The remainder of this article is organized as follows. Section 2 discusses deep learning architectures that have been recently introduced to solve contemporary challenges in vision and speech domain. Section 3 provides a comprehensive discussion of real-world and commercial application cases for the technology. Section 4 discusses state-of-the-art results in implementing these sophisticated algorithms in resource-constrained hardware environments. This section also highlights the prospects of 'smart' applications in mobile devices. Section 5 discusses several successful and emerging applications of neural networks in state-of-the-art IS. Section 6 elaborates potential developments and challenges in the future for IS. Finally, Section 7 concludes with a summary of the key observations in this article.

## 2. Design and Architecture of Neural Networks for Deep Learning

An ANN consists of multiple levels of nonlinear modules arranged hierarchically in layers. This design is inspired by the hierarchical information processing observed in the primate visual system [30, 31]. Such hierarchical arrangements enable deep models to learn meaningful features at different levels of abstraction. Several successful hierarchical ANNs known as deep neural networks (DNNs) are proposed in the literature [32]. Few examples include convolutional neural networks [33], deep belief networks [1], and stacked auto-encoders [34], generative adversarial networks [35], variational autoencoders [36], flow models [37], recurrent neural networks [38], and attention bases models [39]. These models extract both simple and complex features similar to the ones witnessed in the hierarchical regions of the primate vision system. Consequently, the models show excellent performance in solving several computer vision tasks, especially complex object recognition [33]. Cichy *et al.* [30] show that DNN models mimic biological brain function. The results from their object recognition experiment suggest a close relationship between the processing stages in a DNN and the processing scheme observed in the human brain. In the next few sections, we discuss the most popular DNN models and their recent evolutions in various vision and speech applications.

### 2.1 Convolutional neural networks

One of the first hierarchical models, known as convolutional neural networks (CNNs/ConvNets) [33, 40], learns hierarchical image patterns at multiple layers using a series of 2D convolutional operations. CNNs are designed to process multidimensional data structured in the form of multiple arrays or tensors. For example, a 2D color image has three color channels represented by three 2D arrays. Typically, CNNs process input data using three basic ideas: local connectivity, shared weights, and pooling that are arranged in a series of connected layers. A simplified CNN architecture is shown in Fig. 1. The first few layers are convolutional and pooling layers. The convolutional operation processes parts of the input data in small localities to take advantage of local data dependency within a signal. The convolutional layers gradually yield more highly abstract representations of the input data in deeper layers of the network. Another aspect of the convolution operation is that filtering is repeated over the data. This maximizes the use of redundant patterns in the data.

While the convolutional layers detect local conjunctions of features from the previous layer, the role of the pooling layer is to aggregate local features into a more global representation. Pooling is performed by sliding a non-overlapping window over the output of the convolutional layer to obtain a "pooled" value for each window. The pooled value is typically the maximum value over each window, however, averaging or other operations can be applied over the window. This helps a network become robust to small shifts and distortions in input data. The convolutional layer ends by vectorizing the multidimensional data prior to feeding them into fully connected neural networks that perform classification using highly abstracted features from the previous layers. The training of all the weights in the CNN architecture, including the image filters and fully connected network weights, is performed by applying a regular backpropagation algorithm commonly known as gradient-descent optimization.

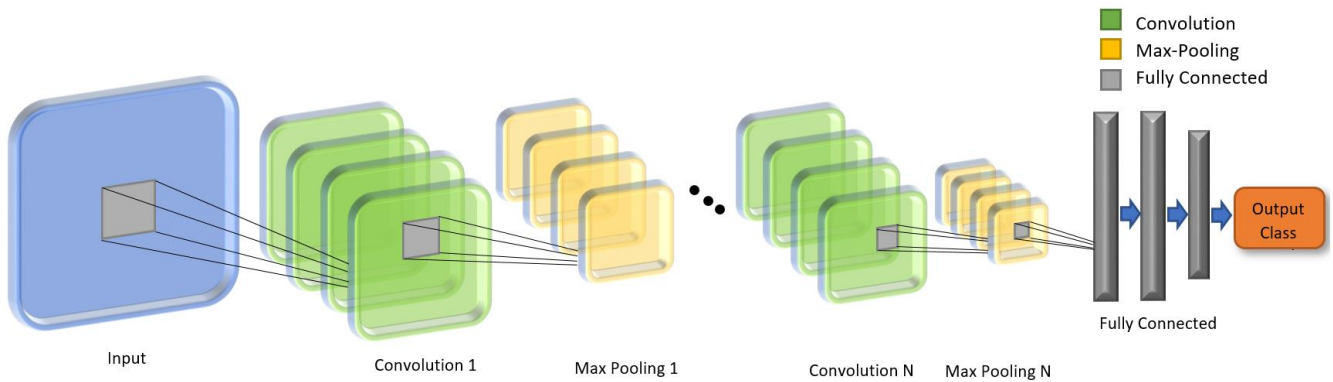

Fig. 1. Generic architecture of Convolutional Neural Network.

### 2.2 Deep generative models and auto-encoders

The hierarchical model of CNN is designed to efficiently learn target-specific features from raw images and videos for vision related applications. However, the major breakthrough of hierarchical models is the introduction of the 'greedy layer-wise' training algorithm for deep belief networks (DBNs) proposed by Hinton *et al.* [28]. A DBN is built in a layer-by-layer fashion by training each learning module known as the restricted Boltzmann machine (RBM) [41]. RBMs are composed of a visible and a hidden layer. The visible layer represents raw data in a less abstract form, and the hidden layer is trained to represent more abstract features by capturing correlations in the visible layer data [41]. Figure 2 (a) shows a standard architecture of a DBN. DBNs are considered hybrid networks that do not support direct end-to-end learning. Consequently, a more efficient architecture, known as deep Boltzmann machines (DBMs) [42], has been introduced. Similar to DBNs, DBMs are structured by stacking layers of RBMs. However, unlike DBNs, the inference procedure of DBMs is bidirectional, allowing them to learn in the presence of more ambiguous and challenging datasets.

The introduction of DBMs has led to the development of the stacked auto-encoder (SAE) [34, 43], which is also formed by stacking multiple layers. Unlike DBNs, SAEs utilize auto-encoders (AE) [44] as the basic learning module. An AE is trained to learn a copy of the input at its output. In doing so, the hidden layer learns an abstract representation of inputs in a compressed form that is known as the encoding units. Figure 2 (b) shows the architecture of an SAE as it gradually learns lower dimensional encoding units at each layer. A greedy layer-wise training algorithm is used to train any of DBN, DBM, or SAE networks, where the parameters of each layer are trained individually by keeping parameters in other layers fixed. After layer-wise training of all layers, also known as pre-training, the hidden layers are stacked together. The entire network with all the stacked layers is then fine-tuned against the target output units to adjust all the parameters for a classification task as illustrated in Fig. 2. DBNs and SAEs have achieved state-of-the-art performance in various vision-related applications such as face verification [45], phone recognition [46], and emotion recognition from image and speech [47, 48]. Moreover, several studies [45, 49] have combined the advantages of different deep learning models to further boost performance in these recognition tasks. For example, Lee *et al.* [49] have shown that combining convolution and weight sharing features of CNNs with the generative architecture of DBNs offers better classification performance on benchmark datasets such as MNIST and Caltech 101 [49]. The hybrid of CNN and DBN models, also known as the CDBN model, enables scaling to problems with large images without requiring an increase in the number of parameters of the network.

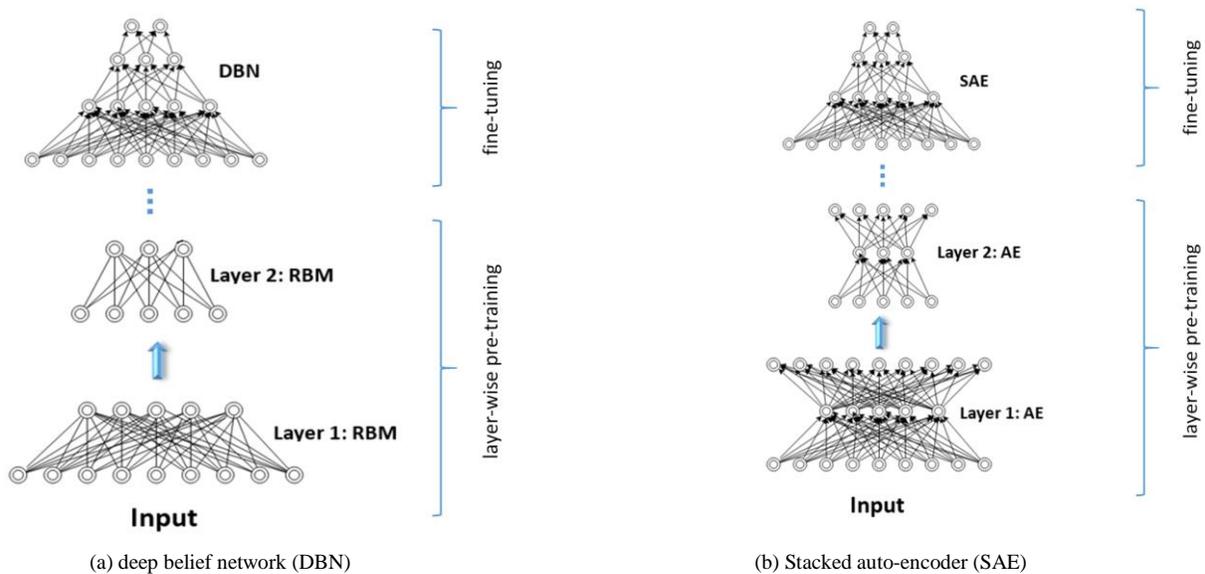

(a) deep belief network (DBN)                    (b) Stacked auto-encoder (SAE)

Fig. 2. A typical architecture including layer-wise pre-training and fine-tuning procedure of (a) deep belief network (DBN); (b) Stacked auto-encoder (SAE).

## 2.3 Variational Autoencoders

Variational autoencoder (VAE) is a generative model that is designed to learn a meaningful latent representation of the input data. The VAE architecture is analogous to an autoencoder, where the deterministic hidden layer is replaced with a parameterizable distribution formulated by variational Bayesian inference. VAE is, therefore, represented by a directed graphical model consisting of an input layer, a probabilistic hidden layer, and an output layer to generate examples that are probabilistically similar to the input class. Kullback Leibler (KL) divergence is used as a constraint between the prior and posterior distribution to achieve a smooth transition in the hidden distributions between different classes. Variation Bayesian inference is used to construct a cost function for the neural network that establishes a connection from the input to hidden layer and then followed by the output layer [36]. The parameterization of hidden layers for several classes can be represented as parameter vectors. Linear combinations of these class-specific vectors can be obtained and used to apply features from different input types into a new output example. VAE has successful applications in image generation [50], motion prediction [51], text generation [52], and expressive speech generation [53].

## 2.4 Generative Adversarial Networks

Generative adversarial network (GAN) is another generative model, that is capable of creating realistic data (typically images) from a given class. A GAN is composed of two competing networks: the generator and the discriminator. The generator aims to generate synthetic images from raw noise input that are as good as real images. The discriminator network has a binary target corresponding to 'fake' or 'real' inputs as it classifies real images against the synthetically generated ones. The entire pipeline of two networks is trained with two alternating goals. One goal is to update the discriminator to improve its classification performance while keeping the generator parameters fixed. The discriminator network yields low cost values when correctly classifying the generator examples as 'fake' against 'real' images. The other goal is to update the generator network by holding fixed parameters for the discriminator. Low cost values for the generator indicates generation of synthetic images that are so real that the discriminator network fails to classify it as 'fake' [35]. Thus, the two networks compete against each other until an optimal point has been reached, which ensures that the fake examples are indistinguishable from real examples. As a generative network, GAN has similar applications to VAE, including image generation [54] and super resolution [55].

The GAN model does not have control over modes of data to be generated. Conditional GAN (CGAN) model alleviates this by adding the ground truth label as an additional parameter to the generator to enforce that the corresponding images are generated. By doing this modification, CGAN allows the GAN model to generate new images from different classes. The generator of the CGAN uses an additional class input to identify the new image type to be generated. The discriminator also has an additional input and only returns 'real' when the input looks real and matches the corresponding input class provided in the generator [56]. The authors in [57] have extended the conditional GAN architecture to construct images from semantic label maps. Bidirectional GAN is concerned with simultaneously learning to generate new images and learning to estimate the latent parameters of existing images [58]. For a given input example, the hidden representation can be extracted. Then the underlying representation can be used to generate a new image of similar semantic quality. The BigBiGan architecture [59] is an improved bidirectional GAN that achieves state-of-the-art results in extraction of image representation and also in image generation tasks.

Despite the popularity and success of GANs, they are frequently plagued by instability in training [60] and subject to underfitting and overfitting [61]. Several studies aimed at improving training stability and performance of GAN. The authors in [62] approach these problems with a weight normalization that they call spectral normalization. Wasserstein GAN (WGAN) is another modification that improves the training of GAN for generating more realistic new example images. The authors in [63] motivate the improvement of GAN with significant theoretical underpinning. The main difference between GAN and WGAN is that instead of providing a binary decision about generated images being 'fake' or 'real', the discriminator network evaluates the generated images using a continuous quality score between 'fake' and 'real'. In [64], the authors consider weight clipping, which is part of WGAN training. Weight clipping is considered as a penalty on the norm of critic gradient, which has shown to improve training stability and image generation quality. In addition to WGAN there are additional works that attempt to improve GAN. For example, least squares generative adversarial networks improve stability and performance [65]. They replace the standard GAN cross-entropy loss with least squares loss to resolve the vanishing gradient problem. Recently, vector quantization is applied to VAE to generate synthetic images of quality rivaling GAN while avoiding the aforementioned problems in training GAN [66].

## 2.5 Flow-Based Models

Flow models construct a decoder that is the exact inverse of the encoder module. This allows exact sampling from the inferred data distribution. In VAE, a distribution parameter vector is extracted by the encoder to define a new distribution that is sampled and decoded to generate an image. In a flow model, given a latent variable, the encoder defines a deterministic transformation into an output image. An early flow model, known as Nonlinear Independent Components Estimation (NICE) [67], is used for generation of images with corrections to corrupt regions of input images, which is known as inpainting. The authors in [37] have extended NICE with several more complex invertible operations, including various types of sampling and masked convolution, to perform image generation. Their proposed model is similar to conditional GAN as it can include additional target class input to constrain the output image class. Another generative model called 'GLOW' uses generative flow with invertible convolutions [68] and is shown capable of generating realistic high-resolution human face images.

## 2.6 Generative Models for Speech

Several related generative models are applied in realistic speech synthesis. WaveNet [69] is an audio generation network based on deep autoregressive models that are used for image generation (e.g. PixelRNN [70]). WaveNet has no recurrent connections, which increases training speed at the cost of increasing the depth of the neural network. In WaveNet, a technique called dilated convolution has been found effective in exponentially increasing the context region with the depth of neural network. WaveNet also utilizes residual connections as described in Section 3.1. Authors in [69] have used conditioning on WaveNet to enable text-to-speech (TTS) generation that yields the state-of-the-art performance when graded by human listeners. Waveglow [71] is another model that combines WaveNet and GLOW for frequency representation of text sequences as input to generate realistic speech. Another model, known as the Speech Enhancement Generative Adversarial Network (SEGAN) [72], uses deep learning and avoids preprocessing speech using spectral domain techniques. The authors use a convolution autoencoder model to input speech and output enhanced speech, and train in a generative adversarial setting. Another work [73] modifies the SEGAN autoencoder model in the context of Wasserstein GAN to perform noise-robust speech enhancement.

## 2.7 Recurrent neural networks

Another variant of neural networks, known as the recurrent neural network (RNN), captures useful temporal patterns in sequential data such as speech to augment recognition performance. An RNN architecture includes hidden layers that retain the memory of past elements of an input sequence. Despite effectiveness in modeling sequential data, RNNs have challenges using the traditional backpropagation technique for training with a sequence of data with larger degrees of separation [38]. The long short-term memory (LSTM) networks alleviate this shortcoming with special hidden units known as "gates" that can effectively control the scale of information to remember or forget in the backpropagation [38]. Bidirectional RNNs [74] consider context from the past as well as the future to process sequential data to improve performance. This, however, can hinder real-time operation as the entire sequence must be available for processing. A modification to LSTM, called Gated Recurrent Unit (GRU) [75], has been introduced in the context of machine translation. The GRU has shown to perform well on translation problems with short sentences. Several variations of LSTM including GRU are compared in [76]. The authors in [76] demonstrate experimentally that, in general, the original LSTM structure is superior for various recognition tasks. LSTM is a powerful model, however, recent advances in attention-based modeling have shown to have better performance than RNN models for sequential and context based information processing [39].

## 2.8 Attention in Neural Networks

The process of attention is an important property of human perception that greatly improves the efficacy of biological vision. The 'attention process' allows humans to selectively focus on particular sections of the visual space to obtain relevant information, avoiding the need to process the entire scene at once. Consequently, the attention provides several advantages in vision processing [77], such as drastic reduction of computational complexity due to the reduction of processing space and improved performance as the objects of importance can always be centralized in the processing space. Additionally, attention models provide noise reduction or filtering by avoiding the processing of irrelevant information in the visual scene and selective fixations over time that allow a contextual representation of the scene without 'clutter'. Hence, the adoption of such methodology for neural network-based vision and speech processing is highly desirable.

Early studies have introduced attention by means of saliency maps (e.g., for mapping of points that may contain important information in an image). A more recent attempt has introduced attention to deep learning models. A seminal study by Larochelle *et al.* [78] models attention in a third-order Boltzmann machine that is able to accumulate information of an overall shape in an image over several fixations. The model is only able to see a small area of an input image, and it learns by gathering information through a sequence of fixations over parts of the image. To learn the sequence of fixations and the overall classification task, the authors in [78] have introduced a hybrid-cost for the Boltzmann machine. This model shows similar performance to deep learning variants that use the whole input image for classification. Another study [79] proposes a two-step system for an attention-based model. First, the whole input image is aggressively downsampled and processed to identify candidate locations that may contain important information. Next, each location is visited by the model in its original resolution. The information collected at each location is aggregated to make the final decision. Similarly, Denil *et al.* [80] have proposed a two-pathway model for object tracking, where one focuses on object recognition and the other pathway works on regulating the attention process.

However, 'learning where and when to attend' is difficult as it is highly dependent on the input and the task. It is also ill-defined in the sense that a particular sequence of fixations cannot be explicitly dictated as ground truth. Due to this challenge, most recent studies on deep learning with attention have employed reinforcement learning (RL) for regulating the attention aspect of the model. Accordingly, a seminal study by Mnih *et al.* [77] builds a reinforcement learning policy on a two-path recurrent deep learning model to simultaneously learn the attention process and the recognition task. Based on similar principles, Gregor *et al.* [81] propose a recurrent architecture for image generation. The proposed architecture uses a selective attention process to trace outlines and generate digits similar to a human. Another study [82] utilizes the selective attention process for image captioning. In this study, the RL based attention process learns the sequence of glimpses through the input image that best describes the scene representation. Conversely, Mansimov *et al.* [83] leverage the RL based selective attention on an image caption to generate new images described in the caption. In this approach, the attention mechanism learns to focus on each word in a sequential manner that is most relevant for image generation. Despite impressive performance in learning selective attention using RL, deep RL still involves additional burdens in developing suitable policy functions that are extremely task specific, and hence, are not generalizable. RL with deep learning also frequently suffers from instability in training.

A different set of studies on designing neural network systems are analogous to the Turing machine architecture that suggests the use of an attention process for interacting with external memory of the overall system. In this approach, the process of attention is implemented using a neural controller and a memory matrix [84]. The attentional focusing allows selectivity of access, which is necessary for memory control [84]. The neural Turing machine work is further explored in [85] considering attention-based global and local focus on an input sequence for machine translation. In [86], an attention mechanism is combined with a bidirectional LSTM network for speech recognition. In [87], the authors, inspired by LSTM for NLP, add a trust gate to augment LSTM for applications in human skeleton-based action recognition. Vaswani *et al.* [39] use an attention module called 'Transformer' to completely replace recurrency in language translation problems. This model is able to achieve improved performance on English-to-German and English-to-French translation. Zhang *et al.* [88] propose self-attention generative adversarial networks (SAGAN) for image generation. A standard convolutional layer can only capture local dependencies in a fixed shape window. Attention mechanism allows the discriminator and generators of the GAN model to operate over larger and arbitrarily shaped context regions [88]. In order to show the growth in deep learning models, Figure 3 summarizes the search results with model names found in the article abstracts as of 2019.

## 2.9 Neural Architecture Search

Neural architecture search (NAS) involves automated selection of the architectural parameters of a neural network. In [89] architectural parameters including CNN filter size, stride, and the number of filters in a given convolutional layer are selected using NAS. Additionally, skip connections (discussed in Section 3.1) are automatically selected to generate densely connected CNN. The method in [89] uses reinforcement learning to train an RNN to generate architectural parameters of a CNN. A more recent method, called the Differential Architecture Search (DARTS) [90], avoids the reinforcement learning paradigm and formulates the problem of parameter selection as a differentiable function that is amenable to gradient descent. The gradient descent formulation improves performance over reinforcement learning and drastically reduces computational time to perform the search. Another work, known as the progressive neural architecture search [91], performs a search over CNN architectures. They begin with a simple structure and progress through a parameter search space toward more complex CNN models. They are also able to reduce the search time and space for the optimal architecture when compared to reinforcement learning methods . They have reported the state-of-the-art performance on the CIFAR-10 image classification dataset. Section 3 elaborates on the contributions of these deep learning models to various vision and speech related applications.

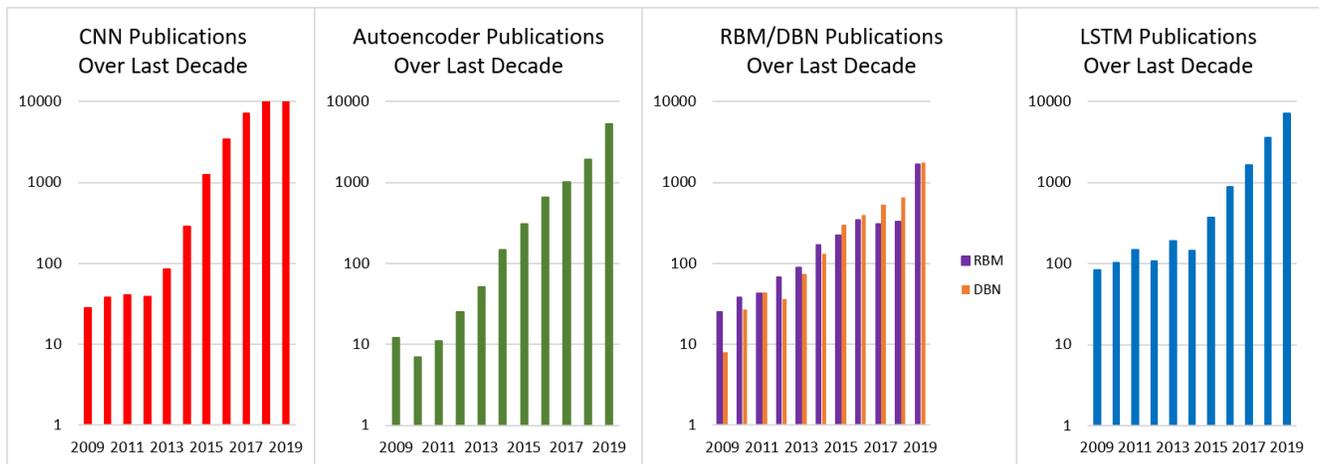

Figure 3. Search for articles showing increasing prominence of deep learning techniques

## 3  DEEP LEARNING IN VISION AND SPEECH PROCESSING

This section discusses the impact of neural networks that are driving the state-of-the-art intelligent vision and speech systems.

### 3.1 Deep learning in computer vision

**Image classification and scene labeling:** The CNN model is first introduced to perform recognition of ten hand-written digits using image examples from the MNIST database. The proposed CNN model has shown significant performance improvement in hand-written digit recognition task compared to earlier state-of-the-art machine learning techniques. Since then CNNs have seen several evolutions and the current versions of CNN are tremendously successful in solving more complex and challenging image recognition tasks [21, 33, 92, 93]. For example, Krizhevsky *et al.* [33] utilize a deep CNN architecture named 'AlexNet' for solving the ImageNet classification challenge [94] to classify 1000 objects from high-resolution natural images. Their proposed CNN architecture has considerably outperformed previous state-of-the-art methods during the earliest attempt with the ImageNet classification challenge. The image recognition performance gradually improved as reported in several publications such as GoogleNet [93], VGGNet [95], ZFNet [96] and ResNet [97], following the initial success of AlexNet. More recently, He *et al.* [98] have extended AlexNet to demonstrate that a carefully trained deep CNN model is able to surpass human-level recognition performance, reported in [94] on the ImageNet dataset. AlexNet [33] and GoogLeNet [93] are two of the pioneering CNN architectures that have significantly improved image classification performance compared to the conventional hand-engineered computer vision models. However, a limitation of these models is the vanishing gradient problem when increasing the number of layers to achieve more depth in learning abstract features. Consequently, a more sophisticated CNN architecture, such as ResNet [97], has been proposed by incorporating "residual block" in the architecture. A residual block is essentially a block with a convolutional operation and a skip connection that are combined into an output. The skip connection directly passes the input with no transformation. This allows the model to achieve very deep structures providing a remedy to the vanishing gradient problem. Densely connected networks are introduced by Huang [99]. They allow forward connections between any two convolutional layers called 'skip' connections. These connections between further-separated layers further reduces vanishing gradient and improves efficiency with reuse of features. Another architecture called squeeze-and-excitation [100] considers channel-wise dependencies in convolutional feature maps. This is performed by calculating the mean statistic for each channel and using this to inform a rescaling of the feature maps. Recently, a technique called EfficientNet [101] is used for scaling of the CNN model. The authors first apply Neural Architecture Search (described in Section 2.8), and then uniformly scale network depth, width, and resolution simultaneously. This method has yielded the state-of-the-art performance in image recognition with an order of magnitude less parameters. The reduction in parameters here also implies faster inference. In Section 4, we extend this discussion of efficient networks for applications in limited resource environments. Scene labeling is another computer vision application that involves assigning target classes to multiple portions of an image based on the local content. Farabet *et al.* [92] have proposed a scene labeling method using a multiscale CNN that yields record accuracies on several scene labeling problem datasets with up to 170 classes. CNNs have also demonstrated the state-of-the-art performance in other computer vision applications, such as in human face, action, expression, and pose recognitions. Table I shows performance error rates of the neural networks described above for image

TABLE I
SUMMARY OF THE SIGNIFICANT STATE-OF-THE-ART CNN IMAGE CLASSIFICATION RESULTS
(*ACTUAL CLASS ERROR WITHIN TOP 5 PREDICTIONS, **PIXEL CLASS ERROR)

| Architecture | Dataset | Error rate |
|---|---|---|
| AlexNet [33] - University of Toronto 2012 | Imagenet (natural images) | 17.0%* |
| GoogLeNet [93] - Google 2014 | Imagenet (natural images) | 6.67%* |
| ResNet [71] - Microsoft 2015 | Imagenet (natural images) | 4.70%* |
| Squeeze & Excitation [100]– Oxford 2018 | Imagenet (natural images) | 2.25%* |
| Multiscale CNN [92] - Farabet et al. 2013 | SIFT/Barcelona (scene labeling) | 32.20%** |

classification.

**Human face, action, and pose recognition:** Human-centric recognitions have long been an active area of research in computer vision. A recent approach in human face recognition is dedicated to improving the cost function of neural networks. The objective of such cost function for face recognition is to maximize interclass variation (facial variations between human individuals) and minimize intraclass variation (facial variations within an individual due to facial expressions). Wang *et al.* [102] have constructed a cost function called large margin cosine loss (LMCL), which achieves the desired variational properties. Using LMCL, their proposed model is able to achieve the state-of-the-art performance on several face recognition benchmarks. Following this work, Deng *et al.* [103] reformulated the cost function for face recognition. Their cost function Additive Angular Margin Loss (ArcFace) is shown to further increase the margin between different face classes. ArcFace is shown to further improve face recognition performance on a large experimental study of 10 datasets. Several CNN-based models are proposed in the literature to perform human action recognition. An architectural feature called temporal pyramid pooling is used in [104] to capture details from every frame in a video and is shown to perform action classification well with a small training set. Another architecture, called the two-stream CNN, analyzes both spatial and temporal context independently and gives competitive results on standard video action benchmarks [95]. CNN architectures that find pose features in an intermediate layer have been used for human action recognition. One of the more successful architectures for action recognition is called R*CNN [105]. This model uses contexts from the scene with human figure data to recognize actions. Action recognition has been performed using a skeletal representation of human individuals instead of RGB video of the entire body posture. Kinect [106] has been widely used to structure illumination of an individual to obtain a 3D skeleton measurement. Kinect skeletons are mapped to color images representing 3D data and used as input in [107] for a ResNet CNN. Tang *et al.* [108] apply reinforcement learning for a graph based CNN (GCNN) that captures structural and temporal information from 3D skeleton input. The authors note that future work may exploit the graph structure in the weight initialization process. Another approach [109] uses raw depth maps and intermediate 3D skeleton features in a multiple channel CNN. A fusion method is applied to the output of different CNN channels, to leverage both modalities. This work improves accuracy on a benchmark with a large number of action classes.

CNNs are used in human pose estimation, for example, Deep Pose [20] is the first CNN application to pose estimation, which has outperformed earlier methods [110]. Deep Pose is a cascaded CNN based pose estimation framework. The cascading allows the model to learn an initial pose estimation based on the full image followed by a CNN based regressor to refine the joint predictions using higher resolution sub-images. Tomson *et al.* [21] propose a 'Spatial Model' which incorporates a CNN architecture with Markov Random Field (MRF) and offers improved results in human pose estimation. Adversarial learning is applied to 2D images in [111] to extend the output pose prediction into 3D space. Furthermore, new sensing techniques allow efficient processing of 3D volumetric data using 3D convolutional networks. For example, in [112], human hand joint locations are estimated in real-time using a volumetric representation of input data and a 3D convolutional network. Another work extends pose estimation to dense pose estimation [113] where the goal is to generate a 3D mesh surface of an individual from 2D images.

**Saliency detection and tracking:** Saliency detection aims to identify regions of an input image that represent an object class of interest. Recent work in saliency detection has proposed the integration of a CNN with an RNN. For example, in [114], RNN is used to refine CNN feature maps by iteratively leveraging deeper contextual information than pure CNN. The work in [115] extends the idea of RNN feature map refinement by introducing multi-path recurrency, which is essentially feedback connection from different depths of CNN. Deep learning has also been applied to detect salient objects in video. One of the recent studies has used 3D CNN [116] to capture the temporal information, while another study [117] incorporates LSTM on top of CNN to capture the temporal information. Recently, Siamese CNN has been proposed to track objects over time in video frames. A Siamese CNN is a two branch CNN that takes two input images. The branches merge in the deep layers to produce an inference on the relation between the two images. In [118, 119], Siamese CNN is used to generate information between adjacent images patches, which is used in tracking objects. Reinforcement learning is another technique that is applied in [120] for tracking biological image features at subpixel level.

**Image generation and inpainting:** Generative models including VAE, GAN and its variants, and flow-based models have applications in image generation and image modification. As mentioned in Section 2.3-2.5, these generative models perform image generation and inpainting including human face image generation. These models are capable of several other applications. In [121], a method called cycle GAN is used for the unpaired image-to-image translation problem. Image-to-image translation would typically involve training on scenes where the input and output domains are given. For example, pairs of pictures of day and night at a location could be a training set. Then given a new image of a location in the day, the network may output a night image. What cycle GAN accomplishes is even more impressive. The training is done without image pairs. So, the day and night images used in training are not from the same locations. The network is then trained to convert day images into night images. Another important GAN application is photo inpainting. When a part of an image is removed or distorted, the network can make a guess of the missing part, for example, face inpainting [122] or natural image inpainting[123]. A recent study has considered partial convolution to perform inpainting with irregularly removed regions [124]. A related application of GAN is semantic image generation. Parts of an image have semantic labels and the goal is to generate an image matching the labeling. The authors in [57] use conditional GAN to generate high-resolution realistic images from semantic maps. A video prediction model based on flow networks has success comparable to VAE in short-period prediction of future frames [125].

TABLE II
COMPARISON OF CONVOLUTIONAL NEURAL NETWORK MODEL CONTRIBUTIONS

| Architecture | Application | Contribution | Limitations |
|---|---|---|---|
| He et al. [98] AlexNet Variant | Image Classification | First human level image classification performance (including fine grained tasks e.g. 100 dog breeds differentiation). Used ReLu generalization and training | Misclassification of image cases that require context |
| Farabet et al. [92] Multiscale CNN | Scene Labeling | Weight sharing at multiple scales to capture context without increasing number of trainable parameters. Global application of graphical model to get consistent labels over the image | Does not apply unsupervised pretraining |
| Wang et al. [104] Temporal Pyramid Pooling CNN | Action Recognition | Temporal pooling for action classification in videos of arbitrary length reduces the chance of overlooking important frames in decision | Challenging similar actions often misclassified |
| Tomson et al. [21] Joint CNN / Graphical Model | Human Pose Estimation | Combining MRF with CNN allows prior belief about joint configurations to impact CNN body part detection | This model works well for constrained set of human poses, general space of human poses remains a challenge |
| Ge et al. [112] 3D CNN | Human Hand Pose Estimation | Volumetric processing of human depth maps of human hands using 3D CNN. 3D reasoning improves occluded finger estimation | Inherently constrained model. Requires clean and presegmented hand regions for pose estimation. The acceptable range of hand joint motion is limited. |

Table II summarizes variants of CNN highlighting their contributions for various computer vision applications with pros and cons. A common theme of CNN models is that these architectures can perform at human level or even better only for simpler tasks. In [98], the authors note that when images require context to explain in image classification, there are more misclassified cases. A similar challenge is observed in human action recognition tasks using visual data or images. Authors in [104] have reported that similar human actions are more challenging to classify using machine algorithms. In [21], the model only works well for a constrained set of human poses. When the classification problems become very difficult such as an arbitrary view or context dependent tasks, the architectures of vision algorithms still have room to improve.

### 3.2 Deep learning in speech recognition

In addition to offering excellent performance in image recognition [21, 33, 92, 93], deep learning models have also shown state-of-the-art performance in speech recognition [126-128]. A significant milestone is achieved in acoustic modeling research with the aid of DBNs at multiple institutions [127]. Following the work in [28], DBNs are trained in layer-wise fashion followed by end-to-end fine-tuning for speech applications as shown in Fig. 2 above. This DBN architecture and training process has been extensively tested on a number of large-vocabulary speech recognition datasets including TIMIT, Bing-Voice-Search speech, Switchboard speech, Google Voice Input speech, YouTube speech, and the English-Broadcast-News speech dataset. DBNs significantly outperform state-of-the-art methods in speech recognition when compared to highly tuned Gaussian mixture model (GMM)-HMM. SAEs likewise are shown to outperform (GMM)-HMM on Cantonese and other speech recognition tasks [43].

RNN has succeeded in improving speech recognition performance because of its ability to learn sequential patterns as seen in speech, language, or time-series data. RNNs have challenges in using traditional backpropagation technique for training such models. This technique has difficulties in using memory to process portions of a sequence with larger degrees of separation [39]. The problem is addressed with the development of long short-term memory (LSTM) networks that use special hidden units known as "gates" to retain memory over longer portions of a sequence [40]. Sak *et al.* [129] first studied the LSTM architecture in speech recognition over a large vocabulary set. Their double-layer deep LSTM is found to be superior to a baseline DBN model. LSTM has been successful in an end-to-end speech learning method, known as Deep-Speech-2 (DS2), for two largely different languages: English and Mandarin Chinese. Other speech recognition studies using an LSTM network have shown significant performance improvement compared to previous state-of-the-art DBN based models. Furthermore, Chien *et al.* [130] performed an extensive experiment with various LSTM architectures for speech recognition and compared the performance with state-of-the-art models. The LSTM model is extended in Xiong *et al.* [131] to bidirectional LSTM. This BLSTM is stacked on top of convolutional layers to improve speech recognition performance. The inclusion of attention enables LSTM models to outperform purely recurrent architectures. An attention mechanism called Listen, Attend, and Spell (LAS) is used to encode, attend, and decode, respectively. This LAS module is used with LSTM to improve speech recognition performance [132]. Using a pretraining technique [133] with attention and LSTM model, speech recognition performance has been improved to a new state-of-the-art level. To summarize key results in speech recognition using DBNs, RNNs (including LSTMs), and attention models, Another memory network based on RNN is proposed by Weston *et al.* [134] to recognize speech content. This memory network stores pieces of information to be able to retrieve the answer related to the inquiry, making it unique and distinctive from standard RNNs and LSTMs. RNN-based models have reached far beyond speech recognition to support natural language processing (NLP). NLP aims to interpret language and semantics from speech or text to perform a variety of intelligent tasks, such as responding to human speech, smart assistants (Siri, Alexa, and Cortana), analyzing sentiment to identify positive or negative attitude towards a situation, processing events or news, and language translation in both speech and texts. Table III summarizes different architectures, datasets used and performance error rates achieved by the state-of-the-art speech recognition models.

TABLE III
SUMMARY OF THE SIGNIFICANT STATE-OF-THE-ART DNN SPEECH RECOGNITION MODELS (*PERPEPLEXITY-SIZE OF MODEL NEEDED FOR OPTIMAL NEXT WORD PREDICTION WITH 10K CLASSES, **WORD ERROR RATE)

| Architecture | Dataset | Error rate |
|---|---|---|
| RNN [126] - <br> FIT, Czech Republic, Johns Hopkins University, 2011 | Penn Corpus <br> (natural language modeling) | 123[*] |
| Autoencoder/DBN [128] - Collaboration, 2012 | English Broadcast News Speech Corpora <br> (spoken word recognition) | 15.5%[**] |
| LSTM [129] - Google, 2014 | Google Voice Search Task (spoken word recognition) | 10.7%[**] |
| Deep LSTM [130] - <br> National Chiao Tung University, 2016 | CHiME 3 Challenge <br> (spoken word recognition) | 8.1%[**] |
| CNN-BLSTM [131] - <br> Microsoft, 2017 | Switchboard <br> (spoken word recognition) | 5.1% |
| Attention (LAS) & LSTM [132] - <br> Google, 2018 | In-house google dictation <br> (spoken word recognition) | 4.1% |
| Attention & LSTM with pretraining [133] - <br> Collaboration, 2018 | LibriSpeech <br> (spoken word recognition) | 3.54% |

TABLE IV
COMPARISON OF RECURRENT NEURAL NETWORK MODEL CONTRIBUTIONS

| Architecture | Application | Contribution | Limitations |
| --- | --- | --- | --- |
| Amodei et al. [159] Gated Recurrent Unit Network | English or Chinese Speech Recognition | Optimized Speech Recognition using Gated Recurrent Units for Speed of Processing achieving near human level results | Deployment requires GPU server |
| Weston et al. [134] Memory Network | Answering questions about simple text stories | Integration of long term memory (readable and writable) component within neural network architecture | Questions and input stories are still rather simple |
| Wu et al. [136] Deep LSTM | Language Translation (e.g. English-to-French) | Multi-layer LSTM with attention mechanism | Especially difficult translation cases and multi-sentence input yet to be tested |
| Karpathy et al. [137] CNN/RNN Fusion | Labeling Images and Image Regions | Use of CNN and RNN together to generate natural language descriptions of images | Fixed image size / requires training CNN and RNN models separately |

Although RNNs/LSTMs are standard in sentiment analysis, authors in [135] have proposed a novel nonlinear architecture of multiple LSTMs to capture sentiments from phrases that constitute different order of the words in natural language. Researchers from Google machine learning [136] have developed a machine-based language translation system that runs Google's popular online translation service. Although this system has been able to reduce average error by 60% compared to the previous system, it suffers from a few limitations. A more efficient translator is used by neural machine translator (NMT) [136] where an entire sentence is input at one time to capture better context and meaning instead of inputting sentences by parts as in traditional methods. More recently, a hybrid approach, combining sequential language patterns from LSTMs and hierarchical learning of images from CNNs, has emerged to describe image content and contexts using natural language descriptions. Karpathy *et al.* [137] introduced this hybrid approach for image captioning to incorporate both visual data and language descriptions to achieve optimal performance in image captioning across several datasets. Table IV summarizes variants of RNN, their pros and cons, and contributions to state-of-the-art speech recognition systems.

Similar to vision tasks, a common theme emerges for RNN models in speech recognition tasks as these architectures can perform at human level or even better for simpler tasks. For both CNNs and RNNs, the architecture is inherently driven by the problem domain. For example: multiscale CNN has been used to gather context for labeling across a scene [92], temporal pooling to understand actions across time [104], MRF graphical modeling on top of CNN to form a prior belief of body poses [21], long term memory component for context retrieval in stories, and CNN fused with RNN to interpret images using language. In [99], the authors note that the question and input stories are rather simple for the neural models to handle. In [101], the authors report that especially difficult translation problems are yet to be successfully addressed in current studies. As tasks become more complex or highly abstract, a more sophisticated intelligent system is required to reach human level performance.

Speech emotion and visual speech recognition are two important topics that have gained recent attention in deep learning literature. Mirsamadi *et al.* [138] have used a deep recurrent network with local attention to automatically learn speech features from audio signals. Their proposed RNN captures a large context region, while the attention focuses on aspects of the speech relevant to emotion detection. This idea is later extended in Chen *et al.* [139] where operation on frequency bank representation of speech signals can be used as inputs into a convolutional layer. This convolutional layer is followed by LSTM and attention layers. Mirsamadi *et al.* have further improved the work of Chen et al. to yield the state-of-the-art performance on Interactive Emotional Dyadic Motion Capture (IEMOCAP) emotion recognition tasks. Another work in [140] applies adversarial auto-encoder for emotion recognition in speech. However, they use heuristic features as network input including spectral and energy features of speech in the IEMOCAP emotion recognition task.

Visual speech recognition involves lip reading of human subjects in video data to generate text captions. Recently, two notable studies have used attention-based networks for this problem. Afouras *et al.* [141] use 3D CNN to capture spatio-temporal information of the face, and a transformer self-attention module guides the network for speech extraction from the extracted convolutional features. Stafylakis *et al.* [142] consider zero-shot keyword spotting, where the phrase is not seen in training and is searched for in a visual speech video. The input video is first fed to a 3D spatial-temporal residual network to capture face information over time. This is followed by attention and LSTM layers to predict the presence of the phrase in the video as well as the moment in time of the phrase. Both studies consider "in the wild" speech recognition or a large breadth of natural sentences in speech.

### 3.3 Datasets for vision and speech applications

Several current datasets have been compiled for state-of-the-art benchmarking of computer vision. ImageNet is a large-scale dataset of annotated images including bounding boxes. This dataset includes over 14 million labeled images spanning more than 20,000 categories [94]. CIFAR-10 is a dataset of smaller images that contain a recognizable object class in low resolution. Each image is only 32x32 pixels, and there are 10 classes with 60,000 images each[143]. Microsoft Common Objects in Context (COCO) provides segmentation of objects in images for benchmarking problems including saliency detection. This dataset includes 2.5 million instances of objects in 328K images [144]. More complex image datasets are now being developed for UAV deployment. Here detection and tracking take place in a highly unconstrained environment. This includes different weather, obstacles, occlusions, and varied camera orientation relative to the flight path. Recently, two large scale datasets were released for benchmarking detection and tracking in UAV applications. The Unmanned Aerial Vehicle Benchmark [145] includes single and multiple bounding boxes for detection and tracking in various flight conditions. An even more ambitious project called Vision Meets Drones [146] gathered a dataset with 2.5 million object annotations for detection and tracking in UAV urban and suburban flight environments.

Speech recognition also has several current datasets for state-of-the-art benchmarking. DARPA commissioned a collaboration between Texas Instruments and MIT (TIMIT) to make a speech transcription dataset. TIMT includes 630 speakers from several American English dialects [147]. VoxCeleb is a more current speech dataset, with 1000 celebrities' voice transcriptions in a more unconstrained or "in the wild" setting [148]. In machine translation, Stanford's natural language processing group has released several public language translation datasets including WMT'15 English-Czech, WMT'14 English-German, and IWSLT'15 English-Vietnamese. The English to Czech and English to German datasets have 15.8 and 4.5 million sentence pairs respectively [149]. CHiME 5 [150] is a speech recognition dataset that contains challenging speech recognition conditions including multiple speaker natural conversations. A dataset called LRS3-TED has been compiled for visual speech recognition [151]. This dataset includes hundreds of hours of TED talk videos with subtitles aligned in time at the resolution of single words. Many other niche datasets can be found on the Kaggle Challenge website free to the public. These datasets include diverse computer vision and speech related problems.

### 3.4 Deep learning in commercial vision and speech applications

In recent years, giant companies such as Google, Facebook, Apple, Microsoft, IBM, and others have adopted deep learning as one of their core areas of research in artificial intelligence (AI). Google Brain [152] focuses on engineering the deep learning methods, such as tweaking CNN-based architectures, to obtain competitive recognition performance in various challenging vision applications using a large number of cluster machines and high-end GPU-based computers. Facebook conducts extensive deep learning research in their Facebook AI Research (FAIR) [153] lab for image recognition and natural language understanding. Many users around the globe are already taking advantage of this recognition system in the Facebook application. Their next milestone is to integrate the deep learning-based NLP approaches to the Facebook system to achieve near human-level performance in understanding language. Recently, Facebook has launched a beta AI assistant called 'M' [154]. 'M' utilizes NLP to support more complex tasks such as purchasing items, arranging delivery of gifts, booking restaurant reservations, and making travel arrangements, or appointments. Microsoft has investigated Cognitive toolkit [155] to show efficient ways to run learning deep models across distributed computers. They have also implemented an automatic speech recognition system achieving human level conversational speech recognition [156]. More recently, they have introduced a deep learning-based speech invoked assistant called Cortana [157]. Baidu has studied deep learning to create massive GPU systems with Infiniband [158] networks. Their speech recognition system named Deep Speech 2 (DS2) [159] has shown remarkably improved performance over its competitors. Baidu is also one of the pioneering research groups to introduce deep learning-based self-driving cars with BMW. Nvidia has invested efforts in developing state-of-the-art GPUs to support more efficient and real-time implementation of complex deep learning models [160]. Their high-end GPUs have led to one of the most powerful end-to-end solutions for self-driving cars. IBM has recently introduced their cognitive system known as Watson [161]. This system incorporates computer vision and speech recognition in a human friendly interface and NLP backend. While traditional computer models have relied on rigid mathematic principles, utilizing software built upon rules and logic, Watson instead relies on what IBM is calling "cognitive computing". The Watson based cognitive computing system has already been proven useful across a range of different applications such as healthcare, marketing, sales, customer service, operations, HR, and finance. Other major tech companies that are actively involved in deep learning research include Apple [162], Amazon [163], Uber [164], and Intel [165]. Figure 4 summarizes publication statistics over the past 10 years searching abstract for 'deep learning', 'computer vision', 'speech recognition', and 'natural language processing' methods applied for computer vision and speech processing.

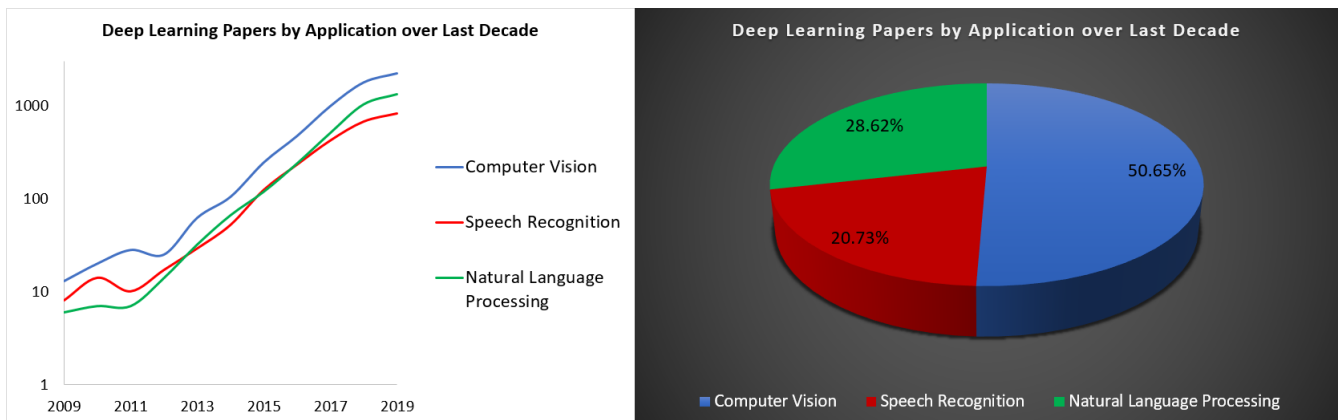

Figure 4. Trends in Deep Learning applications in the literature over the last decade

Although deep learning has revolutionized today's intelligent systems with the aid of computational resources, its applications in more personalized settings, such as in embedded and mobile hardware systems, is another challenge that has led to an active area of research. This challenge is due to the extensive requirement of high-powered and dedicated hardware for executing the most robust and sophisticated deep learning algorithms. Consequently, there is a growing need for developing more efficient, yet robust deep models in resource restricted hardware environments. The next sections summarize some recent advances to develop highly efficient deep models that are compatible with mobile hardware systems.

## 4    VISION AND SPEECH ON RESOURCE RESTRICTED HARDWARE PLATFORMS

The success of future vision and speech systems depends on accessibility and adaptability to a variety of platforms that eventually drive the prospect of commercialization. While some platforms are intended for public and personal usage, there are other commercial, industrial, and online-based platforms - all of which require seamless integration of IS. However, state-of-the-art deep learning models have challenges in adapting to embedded hardware due to large memory footprint, high computational complexity, and high-power consumption. This has led to the research of improving system performance in compact architectures to enable deployment in resource restricted platforms. The following sections highlight some of the major research efforts in integrating sophisticated algorithms in resource restricted user platforms.

### 4.1 Speech recognition on mobile platforms

Handheld devices such as smartphones and tablets are ubiquitous in modern life. Hence, a large effort in developing intelligent systems is dedicated to mobile platforms with a view to reaching out to billions of mobile users around the world. Speech recognition has been a pioneering application in developing smart mobile assistants. The voice input of a mobile user is first interpreted using a speech recognition algorithm. The answer is then retrieved by an online search. The retrieved information is then spoken out by the virtual mobile assistant. Major technology companies, such as Google [166], have enabled voice-based content search on Android devices and a similar voice-based virtual assistant, known as Siri, is also available with Apple's iOS devices. This intelligent application provides mobile users with a fast and convenient hands-free feature to retrieve information.

However, mobile devices, like other embedded systems, have computational limitations and issues related to power consumption and battery life. Therefore, mobile devices usually send input requests to a remote server to process and send the information back to the device. This further brings in issues related to latency due to wireless network quality while connecting to the server. As an example, Keyword spotting (KWS) [167] detects a set of previously defined keywords from speech data to enable hands-free features in mobile devices. The authors in [167] have proposed a low-latency keyword detection method for mobile users using a deep learning-based technique and termed it as 'deep KWS'. The deep KWS method has not only been proven suitable for low-powered embedded systems but also has outperformed the baseline Hidden Markov Models for both noisy and noise-free audio data. The deep KWS uses a fully connected DNN with transfer learning [167] based on speech recognition. The network is further optimized for KWS with end-to-end fine-tuning using stochastic gradient descent. Sainath *et al.* [168] have introduced a similarly small footprint KWS system based on CNNs. Their proposed CNN uses fewer parameters than a standard DNN model, which makes the proposed system more attractive for platforms with resource constraints. Chen *et al.* [169] in another study propose the use of LSTM for the KWS task. The inherent recurrent connections in LSTM can make the KWS task suitable for resource restricted platforms by improving computational efficiency. To support this, the authors further show that the proposed LSTM outperforms a typical DNN-based KWS method. A typical framework for deep learning based KWS system is shown in Fig. 5.

Similar to KWS systems, automatic speech recognition (ASR) [170] has become increasingly popular with mobile devices as it alleviates the need for tedious typing on small mobile devices. Google provides ASR-based search services [166] on Android, iOS, and Chrome platforms, and Apple iOS devices, which are equipped with a conversational assistant named Siri. Mobile users can also type texts or emails by speech on both Android and iOS devices [171]. However, ASR service is contingent on the availability of cellular mobile network since the recognition task is performed on a remote server. This is a limitation since mobile network strength can be low, intermittent, or even absent at places. Therefore, developing an accurate speech recognition system in real-time, embedded on standalone modern mobile devices, is still an active area of research.

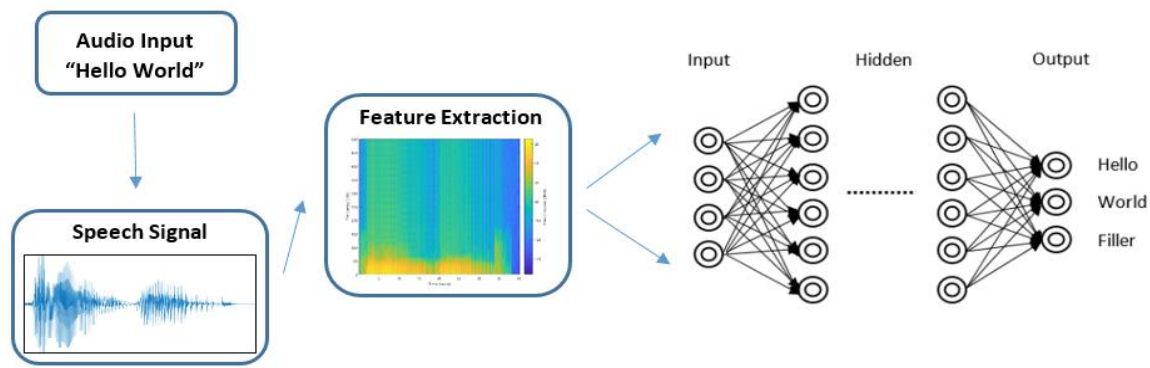

Fig. 5. Generalized framework of a keyword spotting (KWS) system that utilizes deep learning.

Consequently, embedded speech recognition systems using DNNs have gained attention. Lei *et al.* [170] have achieved substantial improvement in ASR performance over traditional gaussian mixture model (GMM) acoustic models even at a much lower footprint and memory requirements. The authors show that a DNN model, with 1.48 million parameters, outperforms the generic GMM-based model while exploiting only 17% of the memory used by GMM. Furthermore, the authors use a language model compression scheme LOUDS [172] to gain a further 60% improvement in the memory footprint for the proposed method. Wang *et al.* [173] propose another compressed DNN-based speech recognition system that is suitable for use in resource restricted platforms. The authors train a standard fully connected DNN model for speech recognition, compress the network using a singular value decomposition method, and then use split vector quantization algorithms to enhance computational efficiency. The authors have achieved a 75% to 80% reduction in memory footprint lowering the memory requirement to a mere 3.2 MB. Additionally, they achieved a 10% to 50% reduction in computational cost with performance comparable to that of the uncompressed version. In [174], the authors show low-rank representation of weight matrices can increase representational power per number of parameters. They also combine this low-rank technique with ensembles of DNN to improve performance on KWS task. Table V summarizes small footprint speech recognition and KWS systems, which are promising for application in resource restricted platforms.

TABLE V

KWS ARCHITECTURES WITH REDUCED COMPUTATIONAL AND MEMORY FOOTPRINT (*RELATIVE IMPROVEMENT OVER COMPARISON NETWORK FROM ROC CURVE, **WER (WORD ERROR RATE), ***RELATIVE FER (FRAME ERROR RATE) OVER COMPARISON NETWORK)

| Compression technique | Memory reduction | Error rate (varied datasets) |
|---|---|---|
| DNN improvement over HMM, 2014 [167] | 2.1M parameters | 45.5% improvement* |
| CNN improvement over DNN, 2015 [168] | 65.5K parameters | 41.1% improvement* |
| Fixed length vector LSTM, 2015 [169] | 152 K parameters | 86% improvement* |
| Split vector quantization, 2015 [173] | 59.1 MB to 3.2MB | 15.8%** |
| Low rank matrices / ensemble training, 2016 [174] | 400 nodes per layer to 100 nodes per layer | -0.174*** |

### 4.2 Computer vision on mobile platforms

Real-time recognition of objects or humans is an extremely desirable feature with handheld devices for convenient authentication, identification, navigational assistance, and when combined with speech recognition, it can even be used as a mobile teaching assistant. Though deep learning has advanced in speech recognition tasks on mobile platforms, image recognition systems are still challenging to deploy in mobile platforms due to the resource constraints.

In a study, Sarkar *et al.* [175] use a deep CNN for face recognition application in mobile platforms for the purpose of user authentication. The authors first identify the disparities in hardware and software between mobile devices and typical workstations in the context of deep learning, such as the unavailability of powerful GPUs and CUDA (an application programming interface by NVIDIA that enables general-purpose processing in GPU) capabilities. The study subsequently proposes a pipeline that leverages AlexNet [33] through transfer learning [176] for feature extraction and then uses a pool of SVM's for scale-invariant classification. The algorithm is evaluated and compared in terms of runtime and face recognition accuracy on several mobile platforms embedded with various Qualcomm Snapdragon CPUs and Adreo GPUs using two standard datasets, UMD-AA [177] and MOBIO [178]. The algorithm has achieved 96% and 88% accuracies with MOBIO and UMD-AA datasets, respectively, with a minimum runtime of 5.7 seconds on the Nexus 6 mobile phone (Qualcomm Snapdragon 805 CPU with Adreno 420 GPU). In another study, Howard *et al.* [179] have introduced a class of efficient CNN models termed 'MobileNets' for mobile and embedded vision processing applications. MobileNets models leverage the depthwise seperability in convolution operation to obtain substantial improvements in efficiency over conventional CNNs. The study also defines two global hyper-parameters that configure the width and depth of the MobileNet architecture to represent a compromise between latency and accuracy in the model performance. Conducting experiments on multiple vision tasks, such as object detection, classification, face recognition, and geo-localization, the authors show approximately seven-fold reduction in trainable parameters using MobileNet at the cost of losing only 1% accuracy when compared to conventional architectures. Su *et al.* [180] have further improved MobileNet by reducing model-level and data-level redundancies that exist in the architecture. Specifically, the authors suggest an iterative pruning strategy [181] to address model-level redundancy, and a quantization strategy [182] to address the data-level redundancy in their proposed architecture. The authors show comparable accuracy of the proposed model with a conventional AlexNet on an ImageNet classification task with just 4% use of trainable parameters and 31% of computational parameters per image inference.

Lane *et al.* [183] have also performed an initial study using two popular deep learning models: CNN and fully connected deep feed-forward networks, to analyze audio and image data on three hardware platforms: Qualcomm Snapdragon 800, Intel Edison, and Nvidia Tegra K1 as these are commonly used in wearable and mobile devices. The study includes extensive analyses on energy consumption, processing time, and memory footprint on these devices when running several state-of-the-art deep models for speech and image recognition applications such as the Deep KWS, DeepEar, ImageNet [33], and SVHN [184] (street-view house number recognition). The study identifies a critical need for optimization of these sophisticated deep models in terms of computational complexity and memory usage for effective deployment in regular mobile platforms.

In another study, Lane *et al.* [185] discuss the feasibility of incorporating deep learning algorithms in mobile sensing for a number of signal and image processing applications. They highlight the limitations that deep models for mobile applications are still implemented on cloud-based systems rather than on standalone mobile devices due to large computational overhead. However, the authors point out that mobile architectures have been advancing in recent years and may soon be able to accommodate complex deep learning methods in devices. The authors subsequently implement a DNN architecture on the Hexagon DSP of a Qualcomm Snapdragon SoC (standard CPU used in mobile phones) and compare its performance with classical machine learning algorithms such as decision tree, SVM, and GMM in processing activity recognition, emotion recognition, and speaker identification. They report increased robustness in performance with acceptable levels of resource use for the proposed DNN implementation in mobile hardware.

### 4.3 Compact, efficient, low power deep learning for lightweight speech and vision processing

As discussed in sections 4.1 and 4.2, hardware constraints pose a major challenge in deploying the most robust deep models in mobile hardware platforms. This has led to a recent research trend that aims to develop compressed but efficient versions of deep models for speech and vision processing. One seminal work in this area is the development of the software platform 'DeepX' by Lane *et al.* [186]. 'DeepX' is based on two resource control algorithms. First, it decomposes large deep architectures into smaller blocks of sub-architectures and then assigns each block to the most efficient local processing unit (CPUs, GPUs, LPUs). Furthermore, the proposed software platform is capable of dynamic decomposition and resource allocation using a resource prediction model [186]. Deploying on two popular mobile platforms, Qualcomm Snapdragon 800 and Nvidia Tegra K1, the authors report impressive improvements in resource use by DeepX for four state-of-the-art deep architectures: AlexNet [33], SpeakerID [187], SVHN [188], and AudioScene in object, face, character, and speaker recognition tasks, respectively [186].

Sindhwani *et al.* [189], on the other hand, propose a memory efficient method using a mathematical framework of structured matrices to represent large dense matrices such as neural network parameters (weight matrices). Structured matrices, such as Toeplitz, Vandermonde, Couchy [190], essentially utilize various parameter sharing mechanisms to represent a $m \times n$ matrix with much less than $mn$ parameters [189]. Authors also show that the use of structured matrices results in substantial improvements in computations, especially in the matrix multiplication operations encountered in deep architectures. The computation time complexity $O(mn)$ is reduced to $O(m \, log(n))$ [189]. This makes both forward computations and backpropagation faster and efficient while training neural networks . The authors test the proposed framework on a deep KWS architecture for mobile speech recognition and compare with other similar small footprint KWS models [168]. The results show that Toeplitz based compression gives the best model computation time, which is 80 times faster than the baseline, at the cost of only 0.4% performance degradation. They also conclude that the compressed model has achieved a 3.6 times reduction in memory footprint compared to the small footprint model proposed in [168].

Han *et al.* [181] propose a neural network-based three-stage compression scheme known as 'deep compression' for reduction of memory footprint. The first stage called pruning [191] essentially removes weak connections in a DNN to obtain a sparse network. The second stage involves trained quantization and weight sharing applied to the pruned network. The third stage uses Huffman coding for lossless data compression in the network. Authors report reduced energy consumption and a significant computing speedup in a comparison between various workstations and mobile hardware platforms. An architecture called ShuffleNet [192] uses two architectural features. Group convolution, introduced in [33], is used with channel shuffle architecture in a novel way to improve the efficiency of convolutional networks. The group convolution improves the speed in processing images and offers comparable performance with reduced model complexity. Table VI summarizes results from different studies for compressed network energy consumption executing AlexNet on a Tegra GPU. Figure 6 summarizes publication statistics over the past 5 years on small footprint analysis of deep learning methods for computer vision, speech processing, and natural language processing in resource restricted hardware platforms.

TABLE VI
COMPRESSED ARCHITECTURE ENERGY AND POWER RUNNING ALEXNET ON A TEGRA GPU

| Compression technique | Execution time | Energy consumption | Implied power consumption |
|---|---|---|---|
| Benchmark study, 2015 [185] | 49.1msec | 232.2mJ | 4.7 W (all layers) |
| Deep X software accelerator, 2016 [186] | 866.7msec (average of 3 trials) | 234.1mJ | 2.7 W (all layers) |
| DNN various techniques, 2016 [181] | 4003.8msec | 5.0mJ | 0.0012W (one layer) |

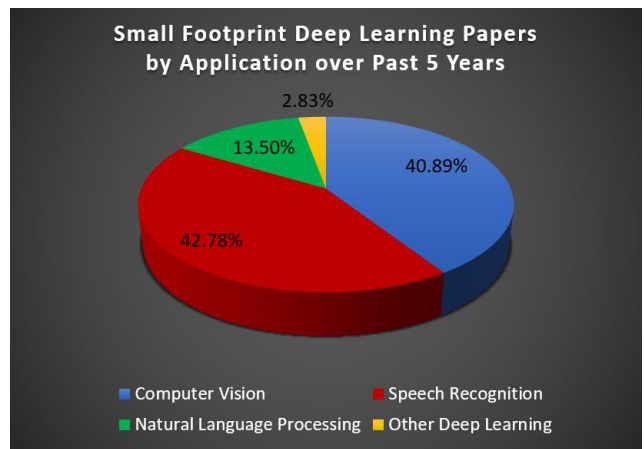

Fig. 6. Publications on small footprint implementations of deep learning in computer and vision and speech processing

## 5 EMERGING APPLICATIONS OF INTELLIGENT VISION AND SPEECH SYSTEMS

We identify three fields of research that are shifting paradigm through recent advances in vision and speech-related frameworks. First, the quantification of human behavior and expressions from visual image and speech offers great potentials in cybernetics, security and surveillance, forensics, quantitative behavioral science, and psychology research [193]. Second, the field of transportation research is rapidly incorporating intelligent vision systems for smart traffic management and self-driving technology. Third, neural networks in medical image analysis show tremendous promise for 'precision medicine'. This represents a vast opportunity to automate clinical measurements, optimize patient outcome predictions, and assist physicians in clinical practice.

### 5.1 Intelligence in behavioral science

The field of behavioral science widely uses human annotations and qualitative screening protocols to study complex patterns in human behavior. These traditional methods are prone to error due to high variability in human rating and qualitative nature in behavioral information processing. Many computer vision studies on human behavior, e.g., facial expression analyses [194], can move across disciplines to revolutionize human behavioral studies with automation and precision.

In behavioral studies, facial expressions and speech are two of the most common means to detect humans' emotional states. Yang *et al.* use quantitative analysis of vocal idiosyncrasy for screening depression severity [23]. Children with neurodevelopmental disorders such as autism are known to have distinctive characteristics in speech and voice [24]. Hence, computational methods for detecting differential speech features and discriminative models [25] can help in the development of future applications to recognize emotion from the voice of children with autism. Recently, deep learning frameworks have been employed to recognize emotion from speech data promising more efficient and sophisticated applications in the future [26, 27, 195].

On the other hand, visual images from videos are used to recognize human behavioral contents [196] such as facial expressions, head motion, human pose, and gestures to support a variety of applications for security, surveillance, and forensics [197-199] and human-computer interactions [19]. The vision-based recognition of facial action units defined by facial action coding system (FACS) [200] has enabled more fine-grain analysis of emotional and physiological patterns beyond prototypical facial expressions such as happiness, fear, or anger. Several commercial applications for real-time and robust facial expression and action unit level analysis have recently appeared in the market with companies such as Noldus, Affectiva, and Emotient. With millions of facial images available for training, state-of-the-art deep learning methods have enabled unprecedented accuracies in these commercially available facial expression recognition applications. These applications are designed to serve a wide range of research studies including classroom engagement analysis [201], consumer preference study in marketing [202], behavioral economics [203], atypical facial expression analysis in neurological disorders [204, 205], and other work in the fields of behavioral science and psychology. The sophistication in face and facial expression analyses may unravel useful markers in diagnosing or differentiating individuals with behavioral or affective dysfunction such as those with autism spectrum disorder [206]. Intelligent systems for human sentiment and expression recognition will play lead roles in developing interactive human-computer systems and smart virtual assistants in the near future.

### 5.2 Intelligence in transportation

Intelligent transportation systems (ITS) cover a broad range of research interests including monitoring driver's inattention [1], providing video-based lane tracking and smart assistance to driving [2], monitoring traffic for surveillance and traffic flow management [3], and more recently tremendous interests in developing self-driving cars [4]. Bojarski *et al.* have recently used deep learning frameworks such as CNN to obtain steering commands from raw images captured by a front-facing camera [5]. The system is designed to operate on highways, without lane markings, and in places with minimal visual guidance. Lane change detection [2, 6] and pedestrian detection [7] have been studied in computer vision and are recently being added as safety features in recent personal vehicles. Similarly, computer vision assisted prediction of traffic characteristics, automatic parking, and congestion detection may significantly ease our efforts in traffic management and safety. Sophisticated deep learning methods, such as LSTM, are being used to predict short term traffic [6], and other deep learning frameworks are being used for predicting traffic speed and flow [8] and for predicting driving behavior [9]. In [10], the authors suggest several aspects of transportation that will be impacted by intelligent systems. Considering multimodal data collection from roadside sensors, RBM will be useful as they are proven to handle multimodal data processing. Considering systems onboard vehicles, CNN can be combined with LSTM to take action in real-time to avoid accidents and improve vehicle efficiency. In line with these research efforts, several car manufacturing companies, such as Audi [207] and Tesla [208], are in active competition for developing next-generation self-driving vehicles with the aid of recent developments in neural network based deep learning techniques. Ride hailing and sharing is another growing domain in transportation. In ride hailing, there is a significant value in predicting pickup demand at different locations to optimize the transportation system and service. CNN has been recently used for location-specific demand of service prediction [209]. Travel time prediction has been performed using CNN and RNN to utilize road network topology and historical trip data [210]. Popular ride sharing services may benefit from recent advances in reinforcement learning. Alabbasi *et al.* have used deep Q-network (a model based on reinforcement learning) along with CNN to develop an optimal vehicle dispatch policy that ultimately improves traffic congestion and emission [211].

## 5.3 Intelligence in medicine

Despite tremendous development in medical imaging techniques, the field of medicine heavily depends on manual annotations and visual assessment of a patient's anatomy and physiology from medical images. Clinically trained human eyes sometimes miss important and subtle markers in medical images resulting in misdiagnosis. Misdiagnosis or even failure to diagnose early can lead to fatal consequences as misdiagnosis is known as the third most common cause of death in the United States [212]. The sophisticated deep learning models with the availability of massive records of multi-institutional imaging databases may ultimately drive the future of precision medicine. Deep learning methods have been successful in medical image segmentation [11], shape and functional measurements of organs [14], disease diagnosis [12], biomarker detection [13], patients' survival prediction from images [213], and many more. Authors in [214] have used a hybrid of LSTM and CNN model to predict patient survival from echocardiographic videos of the heart motion, which has shown a prediction accuracy superior to that of trained cardiologists. Advances in deep neural networks have shown tremendous potential in almost all areas of medical imaging such as ophthalmology [215], dental radiography [216], skin cancer imaging [217], brain imaging [218], cardiac imaging [219, 220], urology [221], lung imaging [222], stroke imaging [223], and so on. In addition to academic research, many commercial companies, including pioneers in medical imaging such as Philips, Siemens, and IBM are investing on large initiatives towards incorporating deep learning methods in intelligent medical image analysis. However, a key challenge remains with the requirement of large ground truth medical imaging data annotated by clinical experts. With commercial initiatives, clinical and multi-institutional collaborations, deep learning-based applications may soon be available in clinical practice.

## 6 LIMITATIONS OF DEEP COMPUTATIONAL MODELS

Despite unprecedented successes of neural networks in recent years, we identify a few specific areas that may greatly impact the future progress of deep learning in intelligent systems. The first area is to develop a robust learning algorithm for deep models that requires a minimal amount of training samples.

## 6.1 Effect of sample size

The current deep learning models require a huge amount of training examples to achieve state-of-the-art performance. However, many application domains may lack such a massive volume of training examples such as in certain medical imaging and behavioral analysis studies. Moreover, prospective acquisition of data may also be expensive in terms of both human and computing resources. The superior performance of deep models comes at the cost of network complexity, which is often hard to optimize and prone to overfitting without a large number of samples to train hundreds and thousands of parameters. Many research studies tend to present over-optimistic performance with deep models without proper validation or proof of generalization across datasets. Some of the solutions such as data augmentation [224, 225], transfer learning [226], and introduction of Bayesian concepts [227, 228] have laid the groundwork for using small data, which we expect to progress over time. The second potential future direction in deep learning research may involve improving the architectures to efficiently handle high dimensional imaging data. In medical imaging, cardiovascular imaging involves time-sampled 3D images of the heart as 4D data. Videos of 3D models and 3D point cloud data involves processing of large volumes of data. The current deep CNN models are primarily designed to handle 2D images. Deep models are often extended to handle 3D volumes by either converting the information to 2D sequences or utilizing dimensionality reduction techniques in the preprocessing stage. This, in turn, results in loss of important information in volume data that may be vital for the analysis. Therefore, a carefully designed deep learning architecture that is capable of efficiently handling raw 3D data similar to their 2D counterparts is highly desirable. Finally, an emerging deep learning research area involves achieving high efficiency for data intensive applications. However, they require careful selection of models, and model parameters to ensure model robustness.

*6.2 Computational burden on mobile platforms*

The computational burden of deep models is one of the major constraints to overcome in making deep models as ubiquitous as the internet of things or to embed in wearable or mobile devices without the connectivity to a remote server. Current state-of-the-art deep learning models utilize an enormous amount of hardware resources, which prohibit deploying them in most practical environments. As discussed in sections 4.1-4.3, we believe that improvements in efficiency and memory footprints may enable the seamless utilization of mobile and wearable devices. An emerging deep learning research area involves achieving real-time learning in memory-constrained applications. Such real-time operation will require careful selection of learning models, model parameterization and sophisticated hardware-software co-design among others.

*6.3 Interpretability of models*

The complexity in network architecture has been a critical factor in providing useful interpretations of model outcomes. In most applications, deep models are used as 'black-box' and optimized using heuristic methods for different tasks. For example, dropout has been introduced to combat model overfitting [227, 229] which is essentially deactivating a number of neurons at random without learning which neurons and weights are truly important to optimize the network performance. More importantly, the importance of input features and the inner working principles are not well understood in deep models. Though there has been some progress to understand the theoretical underpinning of these networks, more work needs to be done.

*6.4 Pitfalls of over-optimism*

In a few applications such as in the game of GO, deep models have outperformed human performance [230] and that has led to the notion that intelligent systems may replace human experts in the future. However, the vision-based intelligent algorithms may not be solely relied on for critical decision-making such as in clinical diagnosis without the supervision of a radiologist, especially where human lives are at stake. While deep neural networks can perform many routine, repetitive, and predictive tasks better than human senses (such as vision) can offer, intelligent machines are unable to master many real-life inherently human level traits such as empathy and many more. Therefore, neural networks are developing intelligent systems that may be better viewed as complementary tools to optimize human performance and decision-making.

## 7   SUMMARY OF SURVEY

This paper systematically reviews the most recent progress in innovating sophisticated intelligent algorithms in vision and speech, their applications, and their limitations in implementation on most popular mobile and embedded devices. The rapid evolution and success of deep learning algorithms has pioneered many new applications and commercial initiatives pertaining to intelligent vision and speech systems, which in turn are improving our daily lives. Despite tremendous success and performance gain of deep learning algorithms, there remain substantial challenges in implementing standalone vision and speech applications on mobile and resource constrained devices. Future research efforts will reach out to billions of mobile phone users with the most sophisticated deep learning-based intelligent systems. From sentiment and emotion recognition to developing self-driving intelligent transportation systems, there is a long list of vision and speech applications that will gradually automate and assist human's visual and auditory perception to a greater scale and precision. With an overview of emerging applications across many disciplines such as behavioral science, psychology, transportation, and medicine, this paper serves as an excellent foundation for researchers, practitioners, and application developers and users.

The key observations for this survey paper are summarized below. First, we provide an overview of different state-of-the-art DNN algorithms and architectures in vision and speech applications. Several variants of CNN models [33, 92-98] are proposed to address critical challenges related to vision-related recognition. Currently, CNN is one of the successful and dynamic areas of research and is dominating state-of-the-art vision systems both in the industry and academia. In addition, we briefly survey several other pioneering DNN architectures, such as DBNs, DBMs, VANs, GANs, VAEs and SAEs in vision and speech recognition applications. RNN models are leading the current speech recognition systems, especially in the emerging applications of NLP. Several revolutionary variants of RNN such as the non-linear structure of LSTM [130, 231] and the hybrid CNN-LSTM architecture [232] have made substantial improvements in the field of intelligent speech recognition and automatic image captioning.

Second, we address several challenges for state-of-the-art neural networks in adapting to compact and mobile platforms. Despite tremendous success in performance, the state-of-the-art intelligent algorithms entail heavy computation, memory usage, and power consumption. Studies on embedded intelligent systems, such as speech recognition and keyword spotting, are focused on adapting the most robust deep language models to resource restricted hardware available in mobile devices. Several studies [167-170, 173] have customized DNN, CNN, and recurrent LSTM architectures with compression and quantization schemes to achieve considerable reductions in memory and computational requirements. Similarly, recent studies on embedded computer vision models suggest lightweight, efficient deep architectures [175, 183, 185] that are capable of real-time performance on existing mobile CPU and GPU hardware. We further identify several studies on developing computational algorithms and software systems [181, 189, 233] that greatly augment the efficiency of contemporary deep models regardless of the recognition task. In addition, we identify the need for further research in developing robust learning algorithms for deep models that can be effectively trained using a minimal amount of training samples. Also, more computationally efficient architecture is expected to emerge to fully incorporate complex 3D/4D imaging data to train the deep models. Moreover, fundamental research in hardware-software co-design is needed to address real-time learning operation for today's memory-constrained cyber and physical systems.

Third, we identify three areas that are undergoing a paradigm shift largely driven by vision and speech-based intelligent systems. The vision or speech-based recognition of human emotion and behavior is revolutionizing a range of disciplines from behavioral science and psychology to consumer research and human-computer interactions. Intelligent applications for driver's assistant and self-driving cars can greatly benefit from vision-based computational systems for future traffic management and driverless autonomous services. Deep neural networks in vision-based intelligent systems are rapidly transforming clinical research with the promise of futuristic precision diagnostic tools. Finally, we highlight three limitations of deep models: pitfalls of using small datasets, hardware constraints in mobile devices, and the danger of over-optimism to replace human experts by intelligent systems.

We hope this comprehensive survey in deep neural networks for vision and speech processing will serve as a key technical resource for future innovations and evolutions in autonomous systems.

ACKNOWLEDGMENT

The authors would like to acknowledge partial funding of this work by the National Science Foundation (NSF) through a grant (Award# ECCS 1310353) and the National Institute of Health (NIH) through a grant (NIBIB/NIH grant# R01 EB020683). Note the views and findings reported in this work completely belong to the authors and not the NSF or NIH.